\documentclass[conference,a4paper]{APSIPA2021}
\usepackage{amsmath}
\usepackage{graphicx}
\usepackage{multirow}
\usepackage{threeparttable}
\usepackage[backend=biber,style=ieee,]{biblatex}
\bibliography{mybib.bib}

\usepackage{geometry}
\usepackage{fancyhdr}

\fancypagestyle{firststyle}{
  \fancyhf{}
  \fancyhead[C]{2024 Asia Pacific Signal and Information Processing Association Annual Summit and Conference (APSIPA ASC)}
}

\begin{document}

\title{Automatic Prompt Generation and Grounding Object Detection for Zero-Shot Image Anomaly Detection}

\author{
\authorblockN{
Tsun-Hin Cheung\authorrefmark{1},
Ka-Chun Fung\authorrefmark{1}\authorrefmark{2},
Songjiang Lai\authorrefmark{1}\authorrefmark{2},
Kwan-Ho Lin\authorrefmark{1}, 
Vincent Ng\authorrefmark{1}, and
Kin-Man Lam\authorrefmark{1}\authorrefmark{2}
}

\authorblockA{
\authorrefmark{1}
Centre for Advances in Reliability and Safety, New Territories, Hong Kong
}

\authorblockA{
\authorrefmark{2}
Department of Electrical and Electronic Engineering, The Hong Kong Polytechnic University, Kowloon, Hong Kong
}
}

\maketitle
\thispagestyle{firststyle}
\pagestyle{fancy}


\begin{abstract}
Identifying defects and anomalies in industrial products is a critical quality control task. Traditional manual inspection methods are slow, subjective, and error-prone. In this work, we propose a novel zero-shot training-free approach for automated industrial image anomaly detection using a multimodal machine learning pipeline, consisting of three foundation models. Our method first uses a large language model, i.e., GPT-3. generate text prompts describing the expected appearances of normal and abnormal products. We then use a grounding object detection model, called Grounding DINO, to locate the product in the image. Finally, we compare the cropped product image patches to the generated prompts using a zero-shot image-text matching model, called CLIP, to identify any anomalies. Our experiments on two datasets of industrial product images, namely MVTec-AD and VisA, demonstrate the effectiveness of this method, achieving high accuracy in detecting various types of defects and anomalies without the need for model training. Our proposed model enables efficient, scalable, and objective quality control in industrial manufacturing settings.
\end{abstract}






\section{Introduction}
\label{sec1}
Ensuring product quality is a major challenge in industrial manufacturing. Identifying defects, flaws, and anomalies in produced goods is crucial to maintaining high standards and avoiding costly recalls or customer dissatisfaction. Traditionally, this quality control process has relied on manual visual inspection by trained human experts. However, this approach is inherently slow, subjective, and prone to human error \cite{sun2024cut}.

Advancements in computer vision and machine learning have enabled the development of automated industrial inspection systems \cite{hridoy2024framework}. These systems can rapidly analyze images of products, detect defects, and classify anomalies in an objective and scalable manner. A key challenge in this domain is capturing the wide variability in the appearance of normal, defect-free products. Conventional approaches using supervised learning often require large, labeled datasets containing both normal and anomalous examples, which can be expensive and time-consuming to collect \cite{saberironaghi2023defect}.

 Recent approaches, such as CLIP \cite{radford2021learning} and WinCLIP \cite{jeong2023winclip}, have achieved promising performance in zero-shot image anomaly detection. However, these CLIP-based methods utilize a set of fixed templates to generate the text prompts of normal and abnormal products, which require human experts and the generated text prompts are not class-specific. This results in limited performance in industrial image anomaly detection because defects are mostly object-specific. Moreover, WinCLIP \cite{jeong2023winclip} applies the sliding windows method to obtain multi-scale features. However, these methods often ignore the different sizes of industrial products.

\begin{table}
\centering
\resizebox{0.48\textwidth}{!}{\begin{tabular}{|l|l|l|} 
\hline
\textbf{Methods} & \textbf{Text Prompt Generation}    & \textbf{Image Region Partition}      \\ 
\hline
CLIP \cite{radford2021learning}             & Template-generated            & Entire image                        \\ 
\hline
WinCLIP \cite{jeong2023winclip}          & Template-generated            & Entire image + sliding windows        \\ 
\hline
\textbf{Ours}     & \textbf{LLM-generated} & \textbf{Entire image + object detector}  \\
\hline
\end{tabular}}
\caption{Comparison of textual prompt generation and image feature extraction of our proposed method to other CLIP-Based image anomaly detection methods.}
\end{table}

To address these two challenges, we propose a novel multimodal approach for industrial image anomaly detection that can address these limitations, as shown in Table I. Our method leverages large language models to automatically generate text descriptions of the expected appearance of the product. These prompts are then used in conjunction with an object detection model and a zero-shot image-text matching model to identify any deviations from the normal state. This approach enables efficient, accurate, and salable quality control without the need for extensive labeled training data. The illustration of our multimodal pipeline is shown in Fig. 1.

\begin{figure}[t]
\centering
\includegraphics[width=\columnwidth]{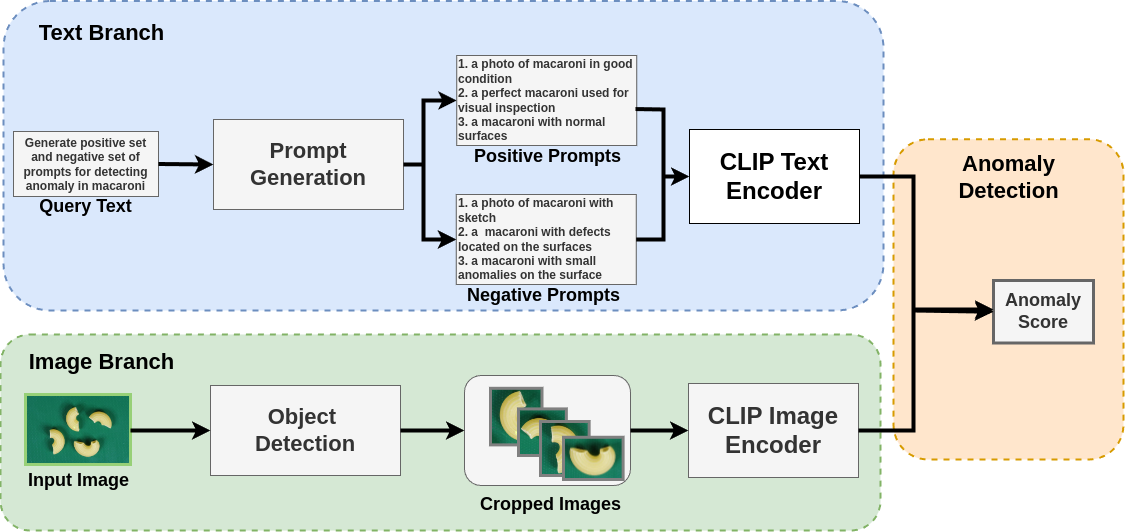}
\caption{Illustration of the prompt generation and object localisation-enhanced CLIP for zero-shot anomaly detection.}\label{fig1}
\end{figure}

Our contributions are summarized as follows:
\begin{itemize}
  \item {We propose to use a language model, such as GPT-3 \cite{brown2020language}, to generate object-specific prompts for describing normal and anomaly industrial images, instead of relying on a set of prompts generated by fixed templates.}
  \item {We propose to use grounding DINO \cite{liu2023grounding} to locate the objects in images, effectively suppressing the background noise and addressing multi-resolution challenges in zero-shot image anomaly detection.}
  \item{Finally, we incorporate the two modules with pretrained CLIP \cite{radford2021learning} to perform image anomaly detection, outperforming the vanilla-CLIP and the current state-of-the-art (SOTA) baseline, i.e., WinCLIP \cite{jeong2023winclip}, in zero-shot settings.}
\end{itemize}

\section{Related Work}

\subsection{Image Anomaly Detection}
Automated image anomaly detection has been a long-standing challenge in computer vision, with applications in various domains such as industrial quality control, medical imaging, and security surveillance. Traditional approaches have often relied on hand-crafted feature extraction and unsupervised anomaly detection techniques, such as one-class support vector machines (OC-SVM) and isolation forests \cite{goyal2020drocc}. Recently, deep learning-based methods have shown promise in addressing the limitations of conventional anomaly detection algorithms. These approaches typically utilize convolutional neural networks (CNNs) or Transformer models to learn representations of normal data and identify deviations from this learned distribution \cite{khan2021spectrogram}. However, a key challenge with these supervised methods is the requirement of labeled anomaly data, which can be costly and time-consuming to obtain in real-world scenarios.

CLIP (Contrastive Language-Image Pretraining) \cite{radford2021learning} is a deep learning model that is pretrained on a large-scale dataset of image-text pairs, enabling it to learn a joint representation of visual and textual data. This learned representation can then be leveraged for zero-shot image anomaly detection, where the model is not explicitly trained on anomalous samples but rather learns to identify deviations from the normal data distribution based on its general understanding of the visual world. WinCLIP \cite{jeong2023winclip} extends the CLIP approach by using a sliding window mechanism instead of analyzing the whole image. In this method, the model is trained to learn a joint representation between image patches and associated textual descriptions. During inference, the model evaluates the aggregated patches and identifies anomalies based on the discrepancy between the visual representation and the normal data distribution. 

In this paper, we also consider the use of CLIP to generate anomaly scores for product images. However, our focus is on improving the quality of text prompts using large language models and enhancing the accurate localisation of products in a CLIP-based pipeline.

\subsection{Prompt Generation with Large Language Models}
The use of large language models, such as GPT-3 \cite{brown2020language} and LLaMA \cite{touvron2023llama}, has revolutionized the field of natural language processing, enabling numerous applications, including text generation, question answering, and sentiment analysis [5, 6].

More recently, researchers have explored the potential of leveraging these powerful language models for tasks beyond pure text processing, such as multimodal learning \cite{liu2024visual} and AI agent reasoning \cite{shen2024hugginggpt}. Our application is slightly different from the language models used for automatic prompt generation \cite{lu2023bounding} in multimodal modeling, where the language model is tasked with producing and enhancing the human written prompts. These language-guided prompts have proven effective in a variety of computer vision tasks, including image classification\ and object detection \cite{he2023unsupervised}. However, the application of prompt generation techniques to industrial image anomaly detection has not been extensively explored and presents a promising area for further research.

\subsection{Grounding Object Detection}
Object detection is a fundamental computer vision task that involves localizing and classifying objects within an image. Traditional object detection methods, such as Fast R-CNN \cite{girshick2015fast} and YOLO \cite{redmon2016you}, have relied on supervised learning approaches that require a large amount of annotated training data. Recent advancements in unsupervised and weakly-supervised object detection have aimed to reduce the need for extensive manual labeling. These include techniques that leverage saliency maps, image-level annotations, or self-supervised learning approaches to learn object representations \cite{zhang2021weakly}. 

Grounding object detection, also known as text-guided object detection, aims to address the limitations of traditional object detection by leveraging natural language descriptions to guide the detection process \cite{sadhu2019zero}. One pioneering work in this area is Referring Expression Comprehension (REC) \cite{kazemzade2014referring}, which focuses on localizing an object in an image based on a natural language description. REC models learn to align visual and textual representations to identify the referred object. Building upon REC, Grounded CLIP \cite{xiao2023clip}  and GroundingDINO \cite{liu2023grounding} extend the grounding concept to generic object detection tasks. These methods utilize the joint representation learning of CLIP \cite{radford2021learning} to associate textual and visual information, enabling zero-shot or few-shot object detection without requiring extensive object-level annotations. In our work, we utilize the zero-shot capability of GroundingDINO to accurately locate the objects, which are then used for anomaly detection.

\section{Methodology}
Our proposed method leverages recent advancements in large language models, object detection, and zero-shot image-text matching to develop an efficient and scalable quality control system. The proposed method consists of three foundation
models. The core idea is to first generate textual prompts describing the expected appearance of the product using language models, i.e., GPT-3 \cite{brown2020language}. Then we use a grounding object detection model, namely Grounding DINO \cite{liu2023grounding} to locate the objects in the input image, and finally compare the located product images to the generated prompts to compute anomaly scores using pretrained CLIP \cite{radford2021learning}.

This multimodal approach allows us to combine the strengths of different deep learning techniques to address the challenges of industrial image anomaly detection. The language model-based prompt generation captures the rich domain knowledge required to define the expected product characteristics, while the object detection and image-text matching components enable robust and accurate anomaly identification.

\subsection{Prompt Generation}
In the text branch, the first step in our method is to generate two sets of textual prompts that describe the expected appearance of the "normal" and "anomaly" product. We utilize the pre-trained GPT-3 large language to generate these prompts.

For the text prompts, we provide the language model with relevant information about the product, including its category. The inputs to the language models to generate the "normal" prompt and "anomaly" prompts are denoted as $x_\text{normal}$ and $x_\text{anomaly}$, respectively. The model then generates two sets of prompts that describe the desired appearance of the product, one for the normal class, and another for the anomaly class, as shown in Equations (1) and (2).

\begin{equation}
\mathbf{P}_\text{normal} = \text{GPT3}(x_\text{normal}),
\end{equation}

\begin{equation}
\mathbf{P}_\text{anomaly} = \text{GPT3}(x_\text{anomaly}),
\end{equation}

where, $\mathbf{P}_\text{normal}$ and $\mathbf{P}_\text{anomaly}$ are two sets of text prompts that describe normal and anomaly products, respectively. 

\subsection{Object Localization}
The next step in our pipeline is to locate the product in the input image. We utilize an object detection model, Grounding DINO to identify the bounding boxes of the products within the image. This step ensures that subsequent anomaly detection is performed on the relevant regions of the image, rather than the entire frame solely.

To locate the product of interest within the input image, we leverage the Grounding DINO object detection model \cite{liu2023grounding}. GroundingDINO is a powerful Transformer-based system that can precisely identify the bounding box coordinates of an object given only its class name as input. We feed the input image $I$ and the class label $c$ of the product into the GroundingDINO model, which then outputs the bounding box coordinates $b = [x, y, w, h]$ that specify the location of the product within the image. This allows us to extract the relevant image patches that contain the product for further analysis. Specifically, we crop the original image $I$ based on the bounding box coordinates to obtain the cropped product image $\mathbf{I}_\text{product}$:

\begin{equation}
\mathbf{I}_\text{object} = \mathbf{I}[y:y+h, x:x+w]
\end{equation}

As multiple bounding boxes can be located in the image, we repeat the above steps to obtain multiple image patches that contains the objects.

\subsection{Zero-Shot Anomaly Detection}
The final step of our method is to detect any anomalies or defects in the cropped product image. We employ a zero-shot image-text matching model, i.e., CLIP \cite{radford2021learning}, to compare the cropped image to the textual prompts generated in the first step.

The image-text matching model is trained to learn a joint embedding space between images and their corresponding textual descriptions. This allows the model to assess the similarity between the input image and the generated prompts, detecting any significant discrepancies that may indicate the presence of an anomaly.

To detect anomalies in product images, we leverage the powerful CLIP,which encodes both text and image data into a shared embedding space, allowing us to compute meaningful similarities between text prompts and image content.

First, we compute the two text embeddings using CLIP - a "normal" prompt that describes the expected appearance of the product, and an "anomaly" prompt that describes potential defects or anomalies we want to detect. The two embedding vectors are denoted as $\mathbf{t}_\text{normal}$ and $\mathbf{t}_\text{anomaly}$, for "normal" and "anomaly" prompts respectively.

To compute the image embedding,  for each cropped product image $\mathbf{I}_\text{object}$,  we use the pretrained CLIP to obtain the average embedding vector of the object patches, denoted as $\mathbf{e}_\text{object}$. We also obtain the image embedding of the whole image as the global feature, denoted as $\mathbf{e}_\text{image}$. Then, the fused feature $\mathbf{e}_\text{fused}$ is obtained by averaging the global feature $\mathbf{e}_\text{image}$ and the local features $\mathbf{e}_\text{object}$.

Following the extraction of the CLIP embeddings for both normal and anomaly prompts, the next step involves determining the anomaly score for each product image. The anomaly score quantifies the deviation of an image from the expected norm, facilitating the identification of potential anomalies. The anomaly score (s) is defined as the following:

\begin{equation}
\text{s} = \frac{\mathbf{e}_\text{fused} \cdot \mathbf{t}_\text{anomaly}}{\mathbf{e}_\text{fused} \cdot \mathbf{t}_\text{anomaly} + \mathbf{e}_\text{fused} \cdot \mathbf{t}_\text{normal}}
\end{equation}

In summary, our anomaly detection method exploits the semantic understanding capabilities of the CLIP model by associating textual prompts with image embeddings, thereby being enable to detect anomalies in product images based on predefined normal and anomaly criteria.

\section{Experiments}
\subsection{Experimental Details}
We evaluate our proposed method on two industrial product image datasets:

MVTec-AD dataset \cite{bergmann2019mvtec}: The MVTec-AD dataset is a widely used benchmark for industrial image anomaly detection. It contains more than 5,000 high-resolution images of 15 different industrial product categories, anootated with various types of anomalies and defects.

VisA dataset \cite{zou2022spot}: The VisA dataset is another industrial product image dataset, containing more than 10,000 images of 20 different product categories. This dataset includes a diverse set of anomalies and defects commonly found in industrial manufacturing environments.

To quantify the performance of our proposed method, we employed two widely used evaluation metrics: Area Under the Receiver Operating Characteristic (AUROC) and Area Under the Precision-Recall Curve (AUPR). AUROC provides a comprehensive measure of the trade-off between the true positive rate and the false positive rate, capturing the overall classification performance. AUPR is a useful metric for evaluating the performance of anomaly detection models, as it takes into account the balance between precision and recall.

\subsection{Comparison to State-of-the-Art Methods}

\begin{table}
\centering
\begin{tabular}{|l|l|l|} 
\hline
\textbf{Methods}~  & \textbf{AUROC}~ & \textbf{AUPR}~   \\ 
\hline
SPADE \cite{cohen2020sub} (1-shot)~    & 0.810~          & 0.906~           \\ 
\hline
PaDiM \cite{defard2021padim} (1-shot)~    & 0.766~          & 0.881~           \\ 
\hline
PathCore \cite{roth2022towards} (1-shot)~ & 0.834~          & 0.922~           \\ 
\hline
WinCLIP \cite{jeong2023winclip} (0-shot)~  & 0.918~          & 0.965~  \\ 
\hline
\textbf{Ours}~     & \textbf{0.932}~ & \textbf{0.966}~           \\
\hline
\end{tabular}
\caption{Comparison of zero-shot image anomaly detection on the MVTec-AD dataset. }
\end{table}

\begin{table}
\centering
\begin{tabular}{|l|l|l|} 
\hline
\textbf{Methods}~  & \textbf{AUROC}~ & \textbf{AUPR}~   \\ 
\hline
SPADE \cite{cohen2020sub} (1-shot)~    & 0.795~          & 0.820~           \\ 
\hline
PaDiM \cite{defard2021padim} (1-shot)~    & 0.628~          & 0.683~           \\ 
\hline
PathCore \cite{roth2022towards} (1-shot)~ & 0.799~          & 0.828~           \\ 
\hline
WinCLIP \cite{jeong2023winclip} (0-shot)~  & 0.781~          & 0.812~           \\ 
\hline
\textbf{Ours}~     & \textbf{0.829}~ & \textbf{0.857}~  \\
\hline
\end{tabular}
\caption{Comparison of zero-shot image anomaly detection on the VisA dataset. }
\end{table}

Table II shows the comparison of our method with several state-of-the-art zero-shot and few-shot image anomaly detection methods on the MVTEC-AD dataset. Our method achieves an AUROC score of 93.2\% and an AUPR score of 96.6\%, outperforming the previous best zero-shot method, WinCLIP \cite{jeong2023winclip}, by a significant margin of over 1 percentage point in both metrics.

The strong performance of our approach on MVTEC-AD demonstrates the effectiveness of our multimodal pipeline in capturing the rich domain knowledge required to accurately define the expected product characteristics and localize anomalies in a zero-shot setting. The language model-based prompt generation, coupled with robust object detection and zero-shot image-text matching components, allows our system to handle a wide variety of defect types, including surface defects, structural anomalies, and missing parts.

Table III presents the results on the VISA dataset, which is another challenging industrial anomaly detection benchmark. Our method achieves an AUROC score of 82.9\% and an AUPR score of 85.7\%, outperforming the current state-of-the-art zero-shot and few-shot methods by a notable margin.

The VISA dataset consists of a diverse set of industrial products, including electronics, automotive parts, and consumer goods, which require a more generalized method for anomaly detection. Our multimodal method is able to adapt to a wider range of product types and defect categories demonstrates its flexibility and robustness. The language model-based prompt generation, combined with the object detection and zero-shot image-text matching, allows our system to effectively capture the unique characteristics of each product and accurately identify anomalies across the diverse VISA dataset.

\subsection{Ablation Studies}

Table IV shows the results of the ablation studies on the MVTEC-AD dataset. We consider two variants of our method: one without the prompt generation component (Ours w/o prompt generation) and another without the object detection component (Ours w/o object detection).

The results demonstrate the importance of both the prompt generation and object detection components in our multimodal pipeline. When removing the object detection, the AUROC and AUPR scores drop by over 1 percentage point, indicating that the language model-based prompt generation is crucial for capturing the rich domain knowledge required for accurate anomaly detection.

\begin{table}
\centering
\begin{tabular}{|l|l|l|} 
\hline
\textbf{Methods}~  & \textbf{AUROC}~ & \textbf{AUPR}~   \\ 
\hline
Ours w/o prompt generation~    & 0.905~          & 0.951~           \\ 
\hline
Ours w/o object detection~    & 0.913~          & 0.953~           \\ 
\hline
\textbf{Ours}~     & \textbf{0.932}~ & \textbf{0.966}~           \\
\hline
\end{tabular}
\caption{Ablation Study on the MVTec-AD dataset. }
\end{table}

\begin{table}
\centering
\begin{tabular}{|l|l|l|} 
\hline
\textbf{Methods}~  & \textbf{AUROC}~ & \textbf{AUPR}~   \\ 
\hline
Ours w/o prompt generation~    & 0.815~          & 0.848~           \\ 
\hline
Ours w/o object detection~    & 0.792~          & 0.821~           \\ 
\hline

\textbf{Ours}~     & \textbf{0.829}~ & \textbf{0.857}~  \\
\hline
\end{tabular}
\caption{Ablation Study on the VisA Dataset. }
\end{table}

Similarly, excluding the prompt generation component leads to a significant performance degradation, with the AUROC and AUPR scores decreasing by more than 2 percentage points. This highlights the importance of object-level features in localizing anomalies in industrial products.

The full multimodal method combines the prompt generation, object detection, and zero-shot image-text matching, achieving the best performance on the MVTEC-AD dataset, with an AUROC of 93.2\% and an AUPR of 96.6\%.

The ablation study results on the VISA dataset are presented in Table V. Similar to the MVTEC-AD analysis, we observe a substantial drop in performance when removing either the prompt generation or the object detection components.

Without the prompt generation, the AUROC and AUPR scores decrease by more than 1 percentage point, demonstrating the value of the language model-based domain knowledge in adapting to the diverse range of products and defect types in the VISA dataset.

The impact of the object detection component is even more pronounced, with the AUROC and AUPR scores dropping by over 3 percentage points when this module is removed. This underscores the critical role of the object-level features in accurately identifying anomalies in the complex VISA dataset. The full multimodal method, combining all the key components, achieves the best performance on the VISA dataset, with an AUROC of 82.9\% and an AUPR of 85.7\%.

The results of the ablation studies on both datasets highlight the complementary nature of the prompt generation, object detection, and zero-shot image-text matching components in our multimodal method. Each module contributes significantly to the overall performance, and the synergistic integration of these elements is crucial for achieving state-of-the-art zero-shot industrial image anomaly detection capabilities.

\section{Conclusion}
In this work, we have proposed a comprehensive approach for industrial anomaly detection that leverages advancements in image anomaly detection, automatic prompt generation, and grounded object localization. By integrating these state-of-the-art techniques, we have developed a robust and versatile method capable of identifying and localizing defects or irregularities in industrial settings.

Our method first employs a deep learning-based anomaly detection model to identify deviations from the learned distribution of normal product samples. This is followed by the use of a large language model to automatically generate descriptive prompts that capture the salient features of the detected anomalies. Finally, we incorporate object detection to localize the regions of interest, providing valuable information for targeted analysis and root cause investigation.

The experimental results on a diverse dataset of industrial products have demonstrated the effectiveness of our method, achieving high accuracy in anomaly detection and precise localization of the detected issues. Furthermore, the language-guided prompts have shown potential for improved interpretability and explainability, which are crucial for industrial applications where the ability to understand and communicate the nature of detected anomalies is of paramount importance.

Moving forward, we aim to further explore the synergies between these technologies, investigating novel architectures and optimization techniques that can enhance the overall performance and robustness of the method. Additionally, we plan to extend the capabilities of our method to handle a wider range of industrial scenarios and explore integration with real-time monitoring and decision-making systems.

Overall, this work demonstrates the promising future of industrial anomaly detection, where the combination of advanced computer vision, natural language processing, and object localization can revolutionize the way we identify and address quality issues in industrial manufacturing and production processes.










\printbibliography

@inproceedings{radford2021learning,
  title={Learning transferable visual models from natural language supervision},
  author={Radford, Alec and Kim, Jong Wook and Hallacy, Chris and Ramesh, Aditya and Goh, Gabriel and Agarwal, Sandhini and Sastry, Girish and Askell, Amanda and Mishkin, Pamela and Clark, Jack and others},
  booktitle={International conference on machine learning},
  pages={8748--8763},
  year={2021},
  organization={PMLR}
}

@inproceedings{jeong2023winclip,
  title={Winclip: Zero-/few-shot anomaly classification and segmentation},
  author={Jeong, Jongheon and Zou, Yang and Kim, Taewan and Zhang, Dongqing and Ravichandran, Avinash and Dabeer, Onkar},
  booktitle={Proceedings of the IEEE/CVF Conference on Computer Vision and Pattern Recognition},
  pages={19606--19616},
  year={2023}
}

@article{brown2020language,
  title={Language models are few-shot learners},
  author={Brown, Tom and Mann, Benjamin and Ryder, Nick and Subbiah, Melanie and Kaplan, Jared D and Dhariwal, Prafulla and Neelakantan, Arvind and Shyam, Pranav and Sastry, Girish and Askell, Amanda and others},
  journal={Advances in neural information processing systems},
  volume={33},
  pages={1877--1901},
  year={2020}
}

@inproceedings{bergmann2019mvtec,
  title={MVTec AD--A comprehensive real-world dataset for unsupervised anomaly detection},
  author={Bergmann, Paul and Fauser, Michael and Sattlegger, David and Steger, Carsten},
  booktitle={Proceedings of the IEEE/CVF conference on computer vision and pattern recognition},
  pages={9592--9600},
  year={2019}
}

@inproceedings{zou2022spot,
  title={Spot-the-difference self-supervised pre-training for anomaly detection and segmentation},
  author={Zou, Yang and Jeong, Jongheon and Pemula, Latha and Zhang, Dongqing and Dabeer, Onkar},
  booktitle={European Conference on Computer Vision},
  pages={392--408},
  year={2022},
  organization={Springer}
}

@article{liu2023grounding,
  title={Grounding dino: Marrying dino with grounded pre-training for open-set object detection},
  author={Liu, Shilong and Zeng, Zhaoyang and Ren, Tianhe and Li, Feng and Zhang, Hao and Yang, Jie and Li, Chunyuan and Yang, Jianwei and Su, Hang and Zhu, Jun and others},
  journal={arXiv preprint arXiv:2303.05499},
  year={2023}
}

@article{hridoy2024framework,
  title={A framework for industrial inspection system using deep learning},
  author={Hridoy, Monowar Wadud and Rahman, Mohammad Mizanur and Sakib, Saadman},
  journal={Annals of Data Science},
  volume={11},
  number={2},
  pages={445--478},
  year={2024},
  publisher={Springer}
}

@article{sun2024cut,
  title={CUT: A Controllable, Universal, and Training-Free Visual Anomaly Generation Framework},
  author={Sun, Han and Cao, Yunkang and Fink, Olga},
  journal={arXiv preprint arXiv:2406.01078},
  year={2024}
}

@inproceedings{goyal2020drocc,
  title={DROCC: Deep robust one-class classification},
  author={Goyal, Sachin and Raghunathan, Aditi and Jain, Moksh and Simhadri, Harsha Vardhan and Jain, Prateek},
  booktitle={International conference on machine learning},
  pages={3711--3721},
  year={2020},
  organization={PMLR}
}

@article{khan2021spectrogram,
  title={A spectrogram image-based network anomaly detection system using deep convolutional neural network},
  author={Khan, Adnan Shahid and Ahmad, Zeeshan and Abdullah, Johari and Ahmad, Farhan},
  journal={IEEE access},
  volume={9},
  pages={87079--87093},
  year={2021},
  publisher={IEEE}
}

@article{saberironaghi2023defect,
  title={Defect detection methods for industrial products using deep learning techniques: A review},
  author={Saberironaghi, Alireza and Ren, Jing and El-Gindy, Moustafa},
  journal={Algorithms},
  volume={16},
  number={2},
  pages={95},
  year={2023},
  publisher={MDPI}
}

@article{touvron2023llama,
  title={Llama: Open and efficient foundation language models},
  author={Touvron, Hugo and Lavril, Thibaut and Izacard, Gautier and Martinet, Xavier and Lachaux, Marie-Anne and Lacroix, Timoth{\'e}e and Rozi{\`e}re, Baptiste and Goyal, Naman and Hambro, Eric and Azhar, Faisal and others},
  journal={arXiv preprint arXiv:2302.13971},
  year={2023}
}

@article{liu2024visual,
  title={Visual instruction tuning},
  author={Liu, Haotian and Li, Chunyuan and Wu, Qingyang and Lee, Yong Jae},
  journal={Advances in neural information processing systems},
  volume={36},
  year={2024}
}

@article{shen2024hugginggpt,
  title={Hugginggpt: Solving ai tasks with chatgpt and its friends in hugging face},
  author={Shen, Yongliang and Song, Kaitao and Tan, Xu and Li, Dongsheng and Lu, Weiming and Zhuang, Yueting},
  journal={Advances in Neural Information Processing Systems},
  volume={36},
  year={2024}
}

@article{lu2023bounding,
  title={Bounding the capabilities of large language models in open text generation with prompt constraints},
  author={Lu, Albert and Zhang, Hongxin and Zhang, Yanzhe and Wang, Xuezhi and Yang, Diyi},
  journal={arXiv preprint arXiv:2302.09185},
  year={2023}
}

@inproceedings{girshick2015fast,
  title={Fast r-cnn},
  author={Girshick, Ross},
  booktitle={Proceedings of the IEEE international conference on computer vision},
  pages={1440--1448},
  year={2015}
}

@inproceedings{redmon2016you,
  title={You only look once: Unified, real-time object detection},
  author={Redmon, Joseph and Divvala, Santosh and Girshick, Ross and Farhadi, Ali},
  booktitle={Proceedings of the IEEE conference on computer vision and pattern recognition},
  pages={779--788},
  year={2016}
}

@article{zhang2021weakly,
  title={Weakly supervised object localization and detection: A survey},
  author={Zhang, Dingwen and Han, Junwei and Cheng, Gong and Yang, Ming-Hsuan},
  journal={IEEE transactions on pattern analysis and machine intelligence},
  volume={44},
  number={9},
  pages={5866--5885},
  year={2021},
  publisher={IEEE}
}

@inproceedings{he2023unsupervised,
  title={Unsupervised prompt tuning for text-driven object detection},
  author={He, Weizhen and Chen, Weijie and Chen, Binbin and Yang, Shicai and Xie, Di and Lin, Luojun and Qi, Donglian and Zhuang, Yueting},
  booktitle={Proceedings of the IEEE/CVF International Conference on Computer Vision},
  pages={2651--2661},
  year={2023}
}

@inproceedings{sadhu2019zero,
  title={Zero-shot grounding of objects from natural language queries},
  author={Sadhu, Arka and Chen, Kan and Nevatia, Ram},
  booktitle={Proceedings of the IEEE/CVF International Conference on Computer Vision},
  pages={4694--4703},
  year={2019}
}

@inproceedings{kazemzade2014referring,
  title={Referring to objects in photographs of natural scenes},
  author={KAZEMZADE, S and Ordonez, V and MATTENV, M and others},
  booktitle={Empirical Methods in Natural Language Processing},
  volume={28},
  pages={787--789},
  year={2014}
}

@article{xiao2023clip,
  title={CLIP-VG: Self-paced Curriculum Adapting of CLIP for Visual Grounding},
  author={Xiao, Linhui and Yang, Xiaoshan and Peng, Fang and Yan, Ming and Wang, Yaowei and Xu, Changsheng},
  journal={IEEE Transactions on Multimedia},
  year={2023},
  publisher={IEEE}
}

@article{cohen2020sub,
  title={Sub-image anomaly detection with deep pyramid correspondences},
  author={Cohen, Niv and Hoshen, Yedid},
  journal={arXiv preprint arXiv:2005.02357},
  year={2020}
}

@inproceedings{defard2021padim,
  title={Padim: a patch distribution modeling framework for anomaly detection and localization},
  author={Defard, Thomas and Setkov, Aleksandr and Loesch, Angelique and Audigier, Romaric},
  booktitle={International Conference on Pattern Recognition},
  pages={475--489},
  year={2021},
  organization={Springer}
}

@inproceedings{roth2022towards,
  title={Towards total recall in industrial anomaly detection},
  author={Roth, Karsten and Pemula, Latha and Zepeda, Joaquin and Sch{\"o}lkopf, Bernhard and Brox, Thomas and Gehler, Peter},
  booktitle={Proceedings of the IEEE/CVF conference on computer vision and pattern recognition},
  pages={14318--14328},
  year={2022}
}

\end{document}